\pdfoutput=1

\documentclass[11pt]{article}

\usepackage[final]{acl}

\usepackage{times}
\usepackage{latexsym}
\usepackage{enumitem}

\usepackage[T1]{fontenc}

\usepackage[utf8]{inputenc}

\usepackage{microtype}

\usepackage{inconsolata}

\usepackage{graphicx}

\usepackage{amsmath}
\usepackage{booktabs}
\usepackage{adjustbox}
\usepackage{multirow}

\usepackage{cleveref}
\usepackage{amsthm}
\newtheorem{definition}{Definition}
\usepackage{graphicx}
\usepackage{subcaption}
\usepackage{tcolorbox}
\tcbuselibrary{skins, breakable, listings}
\usepackage{listings}
\usepackage{amsfonts}
\usepackage{colortbl}
\usepackage{pifont}     %

\usepackage{marvosym}
\usepackage{fontawesome5}
\usepackage{simpleicons}
\usepackage{twemojis}

\definecolor{mine_font}{RGB}{0, 128, 0}
\definecolor{tea_green}{RGB}{214, 234, 193}
\definecolor{hint_green}{RGB}{226,246,209}
\definecolor{Madang}{RGB}{190,235,159}
\definecolor{yellow_green}{RGB}{198,222,119}
\definecolor{link_water}{RGB}{221, 232, 250}
\definecolor{celestial_blue}{RGB}{52, 152, 219}
\definecolor{shakespeare}{RGB}{85, 154, 193}
\definecolor{buttermilk}{RGB}{255,242,174}
\definecolor{chardonnay}{RGB}{250,196,114}
\definecolor{rajah}{RGB}{253,180,98}
\definecolor{fog}{RGB}{213, 193, 234}
\definecolor{melon}{RGB}{254,191,181}
\definecolor{sundown}{RGB}{249, 180, 181}
\definecolor{mona_lisa}{RGB}{246,152,134}
\definecolor{salmon}{RGB}{242,131,107}
\definecolor{blue_x}{RGB}{142, 207, 201}
\definecolor{orange_x}{RGB}{255, 190, 122}

\definecolor{saltpan}{RGB}{238, 243, 232}
\definecolor{aqua_spring}{RGB}{232, 243, 232}
\definecolor{tea_green}{RGB}{214, 234, 193}
\definecolor{Madang}{RGB}{190,235,159}
\definecolor{fringy_flower}{RGB}{194, 234, 193}
\definecolor{aero_blue}{RGB}{193, 234, 213}
\definecolor{pixie_green}{RGB}{183,214,170}
\definecolor{french_pass}{RGB}{195,232,246}
\definecolor{ice_cold}{RGB}{169,232,220}
\definecolor{pale_turquoise}{RGB}{172,240,242}
\definecolor{cruise}{RGB}{179,226,205}
\definecolor{sail}{RGB}{163,205,235}
\definecolor{spindle}{RGB}{179,205,227}
\definecolor{link_water}{RGB}{221, 232, 250}
\definecolor{periwinkle}{RGB}{203,213,232}
\definecolor{zanah}{RGB}{220, 233, 213}
\definecolor{frostee}{RGB}{217, 231, 214}
\definecolor{opal}{RGB}{199, 221, 211}
\definecolor{jet_stream}{RGB}{188, 214, 210}
\definecolor{skeptic}{RGB}{153, 187, 167}
\definecolor{hint_green}{RGB}{226,246,209}
\definecolor{snow_flurry}{RGB}{230,245,201}
\definecolor{surf_crest}{RGB}{205,230,208}
\definecolor{yellow_green}{RGB}{198,222,119}
\definecolor{cream}{RGB}{255,255,204}
\definecolor{pale_prim}{RGB}{255,255,179}
\definecolor{spring_sun}{RGB}{242,243,195}
\definecolor{portafino}{RGB}{245,237,160}
\definecolor{buttermilk}{RGB}{255,242,174}
\definecolor{cream_brulee}{RGB}{255, 229, 151}
\definecolor{dairy_cream}{RGB}{254,226,189}
\definecolor{champagne}{RGB}{254,217,166}
\definecolor{chardonnay}{RGB}{250,196,114}
\definecolor{manhattan}{RGB}{226,180,125}
\definecolor{rajah}{RGB}{253,180,98}
\definecolor{early_dawn}{RGB}{252,243,218}
\definecolor{egg_shell}{RGB}{238, 234, 215}
\definecolor{selago}{RGB}{243, 232, 243}
\definecolor{quartz}{RGB}{219,223,238}
\definecolor{fog}{RGB}{213, 193, 234}
\definecolor{languid_lavender}{RGB}{222,203,228}
\definecolor{watusi}{RGB}{254,221,207}
\definecolor{coral_andy}{RGB}{243,204,205}
\definecolor{cosmos}{RGB}{248,209,210}
\definecolor{melon}{RGB}{254,191,181}
\definecolor{azalea}{RGB}{234, 193, 194}
\definecolor{beauty_bush}{RGB}{235, 185, 179}
\definecolor{sundown}{RGB}{249, 180, 181}
\definecolor{mona_lisa}{RGB}{246,152,134}
\definecolor{salmon}{RGB}{242,131,107}

\definecolor{summer_sky}{RGB}{58, 151, 233}
\definecolor{chateau_green}{RGB}{72, 179, 96}
\definecolor{matisse}{RGB}{25, 104, 167}
\definecolor{allports}{RGB}{31, 106, 125}
\definecolor{sun_shade}{RGB}{255, 144, 68}
\definecolor{flamingo}{RGB}{237, 88, 85}
\definecolor{studio}{RGB}{128, 91, 160}

\definecolor{maya_blue}{RGB}{102, 204, 255}
\definecolor{feijoa}{RGB}{178,223,138}
\definecolor{sushi}{RGB}{117, 168, 47}
\definecolor{norway}{RGB}{158, 194, 132}
\definecolor{japanese_laurel}{RGB}{53, 116, 40}
\definecolor{see_green}{RGB}{161,228,195}
\definecolor{monte_carlo}{RGB}{135,204,194}
\definecolor{granny_smith_apple}{RGB}{150,214,150}
\definecolor{moss_green}{RGB}{170,216,176}
\definecolor{chateau_green}{RGB}{72, 179, 96}
\definecolor{opal}{RGB}{164,207,190}
\definecolor{acapulco}{RGB}{117, 170, 148}
\definecolor{viridian}{RGB}{55, 137, 122}
\definecolor{amazon}{RGB}{56, 123, 84}
\definecolor{asparagus}{RGB}{123, 160, 91}
\definecolor{fruit_salad}{RGB}{91, 160, 94}
\definecolor{puerto_rico}{RGB}{72, 179, 150}
\definecolor{mountain_meadow}{RGB}{0, 163, 136}
\definecolor{matisse}{RGB}{25, 104, 167}
\definecolor{allports}{RGB}{31, 106, 125}
\definecolor{astral}{RGB}{55, 111, 137}
\definecolor{spring_leaves}{RGB}{46, 83, 117}
\definecolor{biscay}{RGB}{44, 62, 80}
\definecolor{midnight}{RGB}{0, 29, 50}
\definecolor{amethyst}{RGB}{153, 102, 204}
\definecolor{studio}{RGB}{128, 91, 160}
\definecolor{tapestry}{RGB}{194, 109, 132}
\definecolor{atomic_tangerine}{RGB}{255, 153, 102}
\definecolor{amber}{RGB}{255, 191, 0}
\definecolor{casablanca}{RGB}{244, 178, 84}
\definecolor{california}{RGB}{233, 140, 58}
\definecolor{tomato}{RGB}{255, 97, 56} 
\definecolor{alizarin}{RGB}{233, 58, 64}

\definecolor{linen}{RGB}{251, 239, 227}
\definecolor{double_pearl_lusta}{RGB}{253, 242, 208}
\definecolor{oasis}{RGB}{253, 242, 208}
\definecolor{milan}{RGB}{255, 254, 169}
\definecolor{texas}{RGB}{245, 232, 123}
\definecolor{maize}{RGB}{249, 212, 156}

\definecolor{turmeric}{RGB}{211, 178, 76}
\definecolor{saffron}{RGB}{249,193,62}
\definecolor{my_sin}{RGB}{255, 176, 59}
\definecolor{tree_poppy}{RGB}{246, 154, 27}
\definecolor{jaffa}{RGB}{240, 131, 58}
\definecolor{crusta}{RGB}{254, 127, 44}
\definecolor{tahiti_gold}{RGB}{223, 102, 36}
\definecolor{outrageous_orange}{RGB}{255, 100, 45}
\definecolor{safety_orange}{RGB}{254, 106, 0}

\definecolor{azalea}{RGB}{251, 196, 196}
\definecolor{oyster_pink}{RGB}{238,206,205} 
\definecolor{coral_candy}{RGB}{242,208,205} 
\definecolor{baby_pink}{RGB}{246, 194, 192}
\definecolor{petite_orchid}{RGB}{223, 157, 155}
\definecolor{apricot}{RGB}{241,140,122}
\definecolor{NY_pink}{RGB}{228,136,113}
\definecolor{carmine_pink}{RGB}{231, 76, 60}
\definecolor{deep_carmine_pink}{RGB}{236, 50, 67}

\definecolor{wewak}{RGB}{244, 143, 150}
\definecolor{light_coral}{RGB}{244, 127, 123}
\definecolor{bittersweet}{RGB}{255,111,105}
\definecolor{carnation}{RGB}{245, 80, 86}
\definecolor{flamingo}{RGB}{237, 88, 85}
\definecolor{sunset_orange}{RGB}{242,89,75}
\definecolor{ku_crimson}{RGB}{243, 0, 25}
\definecolor{amaranth}{RGB}{234,46,73}
\definecolor{valencia}{RGB}{214, 87, 70}
\definecolor{chilean_fire}{RGB}{215, 87, 44}
\definecolor{mexican_red}{RGB}{170, 41, 37}

\definecolor{napa}{RGB}{163, 154, 137}

\definecolor{athens_gray}{RGB}{236, 240, 241}
\definecolor{gallery}{RGB}{240,240,240}
\definecolor{mercury}{RGB}{230,230,230}
\definecolor{platinum}{RGB}{228,228,228}
\definecolor{silver}{RGB}{191,191,191}
\definecolor{aluminum}{RGB}{153,153,153}
\definecolor{ship_gray}{RGB}{77,77,77}
\definecolor{tuatara}{RGB}{67, 67, 67}

\definecolor{malibu}{RGB}{110, 180, 240}
\definecolor{celestial_blue}{RGB}{52, 152, 219}
\definecolor{curious_blue}{RGB}{41, 128, 185}
\definecolor{french_blue}{RGB}{0, 112, 182}
\definecolor{matisse}{RGB}{25, 104, 167}
\definecolor{shakespeare}{RGB}{85, 154, 193}
\definecolor{seagull}{RGB}{128,177,211}
\definecolor{jelly_bean}{RGB}{45, 126, 150}
\definecolor{venice_blue}{RGB}{87, 135, 105}
\definecolor{boston_blue}{RGB}{68, 147, 161}

\definecolor{turquoise}{RGB}{41,217,194}
\definecolor{java}{RGB}{2,190,196}
\definecolor{riptide}{RGB}{141,211,199}
\definecolor{mountain_meadow}{RGB}{0, 163, 136}
\definecolor{free_speech_aquamarine}{RGB}{0, 156, 114}

\definecolor{cosmic_latte}{RGB}{222, 247, 229}
\definecolor{chinook}{RGB}{163, 232, 178}
\definecolor{padua}{RGB}{121, 189, 143}
\definecolor{ocean_green}{RGB}{79, 176, 112}
\definecolor{pastel_green}{RGB}{107, 227, 135}
\definecolor{chateau_green}{RGB}{69, 191, 85}
\definecolor{RoyalBlue}{RGB}{69, 191, 85}
\definecolor{pigment_green}{RGB}{0, 175, 79}
\definecolor{fern}{RGB}{101,197,117}
\definecolor{killarney}{RGB}{56, 113, 66}

\newcommand{\unvicon}{\scalebox{0.8}{\tiny\faIcon{university}}}
\newcommand{\shieldicon}{\scalebox{0.8}{\tiny\faIcon{shield-alt}}}
\newcommand{\meituanicon}{\scalebox{0.9}{\tiny\faIcon{running}}}

\newcommand{\numinc}[2]{%
  \begin{tikzpicture}[baseline]
    \pgfmathsetmacro{\p}{min(130,130*#2/0.10)} %
    \fill[monte_carlo!\p!white, rounded corners=1] (-0.6em,-0.3em) rectangle (2.8em,1em);
    \node[inner sep=0pt] at (1em,0.7ex) {#1};
  \end{tikzpicture}%
}

\newcommand{\colorbarverticalinc}[1][5.2]{%
  \begin{tikzpicture}[x=1cm,y=1cm,baseline=(c.base)]
    \coordinate (c) at (0,0);
    \def\H{#1}
    \shade[top color=white, bottom color=monte_carlo!150, rounded corners=0.6]
      (0,0) rectangle (0.28,\H);
    \foreach \p/\lab in {0/0,0.2/2,0.4/4,0.6/6,0.8/8,1/10}{
      \draw[white,semithick] (0,\H-\p*\H) -- (0.05,\H-\p*\H);
      \node[right] at (0.33,\H-\p*\H) {\scriptsize \lab};
    }
    \node[rotate=90, anchor=south] at (1.10,0.5*\H) {\scriptsize \textbf{Improvement (\%)}};
  \end{tikzpicture}%
}

\title{Too Consistent to Detect: A Study of Self-Consistent Errors in LLMs}

\author{Hexiang Tan$^{\shieldicon\unvicon}$  \hspace{0.5em} Fei Sun\textsuperscript{\shieldicon\,\tiny\textcolor{matisse}{\faIcon[regular]{envelope}}} \hspace{0.5em}  Sha Liu$^{\unvicon}$ \hspace{0.5em} Du Su$^{\shieldicon}$ \hspace{0.5em}  Qi Cao$^{\shieldicon}$ \hspace{0.5em} \textbf{Xin Chen}$^{\meituanicon}$ \\ 
  \textbf{Jingang Wang}$^{\meituanicon}$ \hspace{0.1em} \textbf{Xunliang Cai}$^{\meituanicon}$ \hspace{0.1em} \textbf{Yuanzhuo Wang}$^{\shieldicon}$  \hspace{0.1em} \textbf{Huawei Shen}$^{\shieldicon\unvicon}$ \hspace{0.1em} \textbf{Xueqi Cheng}$^{\shieldicon\unvicon}$\\
  $^{\shieldicon}$State Key Laboratory of AI Safety, Institute of Computing Technology, CAS\\
  $^{\unvicon}$University of Chinese Academy of Sciences \hspace{2.5em} $^{\meituanicon}$Meituan\\
 \tt{tanhexiang21s@ict.ac.cn \,\,\, \textsuperscript{\tiny\textcolor{matisse}{\faIcon[regular]{envelope}}}sunfei@ict.ac.cn}}

\begin{document}
\maketitle

\renewcommand*{\thefootnote}{\tiny\textcolor{matisse}{\faIcon[regular]{envelope}}}
\footnotetext{\textit{Corresponding author}: Fei Sun (\href{sunfei@ict.ac.cn}{sunfei@ict.ac.cn})}
\renewcommand*{\thefootnote}{\tiny\textcolor{matisse}{\faIcon{users}}} %
\footnotetext{\textit{Author contributions}: Hexiang Tan proposed the idea and conducted the main experiments; Fei Sun led the design and supervision; Sha Liu performed supporting experiments.} 
\renewcommand*{\thefootnote}{\arabic{footnote}}

\begin{abstract}
As large language models (LLMs) often generate plausible but incorrect content, error detection has become increasingly critical to ensure truthfulness.
However, existing detection methods often overlook a critical problem we term as \textbf{self-consistent error}, where LLMs repeatedly generate the \textit{same} incorrect response across multiple stochastic samples.
This work formally defines self-consistent errors and evaluates mainstream detection methods on them.
Our investigation reveals two key findings: 
(1) Unlike inconsistent errors, whose frequency diminishes significantly as the LLM scale increases, the frequency of self-consistent errors remains stable or even increases.
(2) All four types of detection methods significantly struggle to detect self-consistent errors.
These findings reveal critical limitations in current detection methods and underscore the need for improvement.
Motivated by the observation that self-consistent errors often differ across LLMs, we propose a simple but effective \textit{cross‑model probe} method that fuses hidden state evidence from an external verifier LLM.
Our method significantly enhances performance on self-consistent errors across three LLM families\footnote{Code Released at \href{https://github.com/Tan-Hexiang/Too-Consistent-to-Detect}{https://github.com/Tan-Hexiang/Too-Consistent-to-Detect}}.

\end{abstract}

\section{Introduction}
As large language models (LLMs) are increasingly deployed in high-stakes applications \citep{chen2024a}, their tendency to generate plausible yet incorrect content raises critical safety concerns.
Therefore, error detection has become essential for ensuring the trustworthiness of LLMs \citep{selfcheckgpt, generating_with_confidence, farquhar2024detecting}.
Numerous error detection methods rely on measuring consistency across multiple samples \citep{selfcheckgpt, generating_with_confidence, semantic_entropy, inside,xue2025verify} under the assumption that consistent outputs are more likely to be correct.
However, this assumption fails to account for a crucial phenomenon we define as ``\textbf{self-consistent error}'', where LLMs \textit{consistently generate semantically equivalent errors across multiple stochastic samples for the same question}, in contrast to ``\textbf{inconsistent error}'', which varies between samples.

\begin{figure}[t]
    \centering
    \begin{adjustbox}{width=0.97\columnwidth}
        \includegraphics{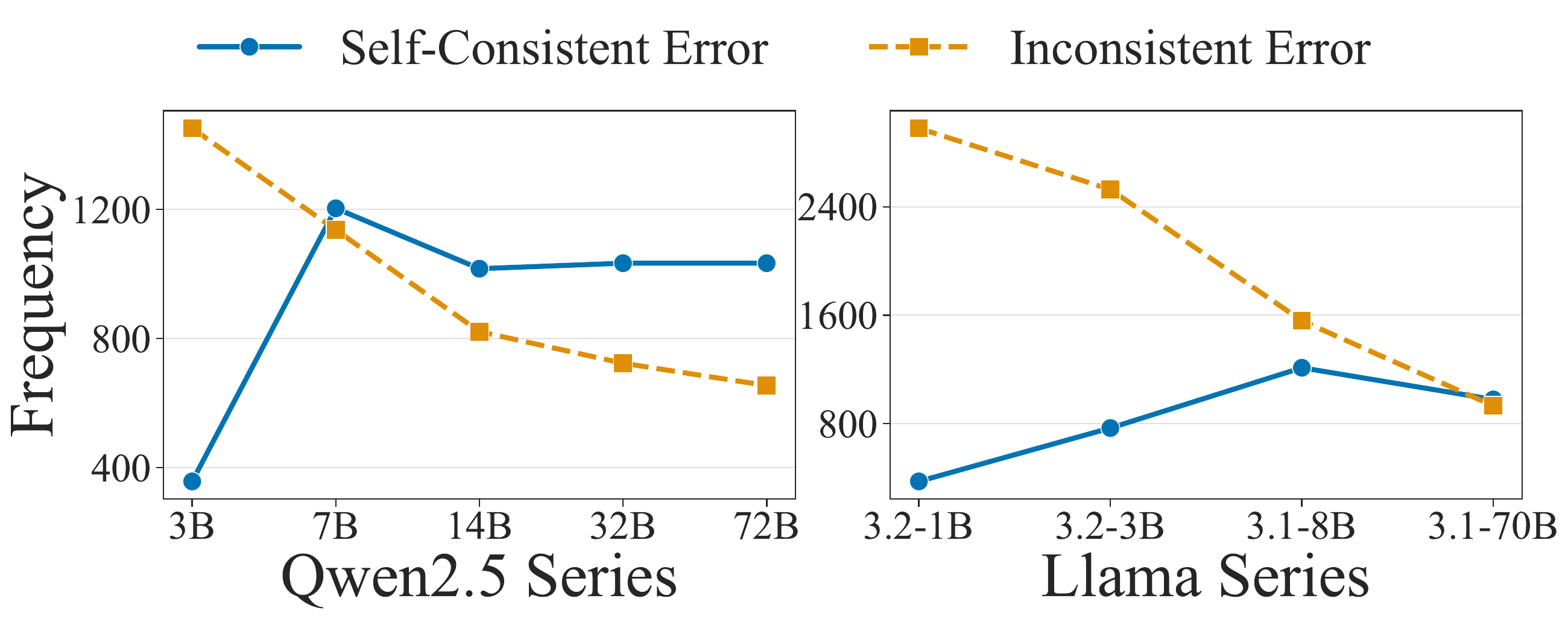}
    \end{adjustbox}
    \caption{Frequency of self-consistent and inconsistent errors across different model scales on SciQ. Inconsistent errors decrease with model size while self-consistent errors remain stable or even slightly increase.}
    \label{fig:intro}
\end{figure}

To demonstrate the importance of self-consistent errors, we analyze their frequency across the SciQ and TriviaQA datasets using nine model scales from the Qwen and Llama series.
Figure \ref{fig:intro} shows that the frequency of self-consistent errors remains stable or even increases with model scale, while inconsistent errors decrease significantly.
This divergence highlights that self-consistent errors remain resistant to scaling, posing a persistent and long-term challenge.
Therefore, detecting self-consistent errors becomes a critical research goal.

This paper systematically evaluates four types of mainstream error detectors on self-consistent errors, including probability methods \citep{duan2024shifting}, prompt-based \citep{kadavath2022languagemodelsmostlyknow,just_ask_for_calibration,can_llm_express}, supervised probe-based \citep{lying, InternalInspector, pollmgraph}, and consistency-based methods.
We find that all methods suffer substantial performance drops on self-consistent errors, in contrast to their strong performance on inconsistent errors.
Consistency-based detectors degrade the most, even falling below random guessing (AUROC $\leq 0.5$). 
Notably, even the strongest supervised probe that accesses the model's hidden states shows significant performance drops, suggesting that the hidden states of an LLM alone cannot provide sufficient signal for detecting self-consistent error.

To improve detecting self-consistent errors, we propose a novel \textit{cross-model probe} based on an observation: self-consistent errors tend to be model-specific and rarely overlap across different LLMs.
Inspired by this, we feed the original model’s response into an external verifier, extract its hidden states, and train a dedicated probe on them.
The verifier-based probe is then integrated with the original probe to produce a unified detection score.
This cross-model perspective compensates for the blind spots of the original model, enabling more reliable detection.
Experiments across three LLM families and two datasets demonstrate that our method achieves substantial improvements in detecting self-consistent errors, offering a promising direction for future detection methods.

\begin{table*}[t]
\centering
\small
\begin{adjustbox}{max width=0.98\linewidth}
    \begin{tabular}{m{0.13\textwidth} m{0.16\textwidth} m{0.13\textwidth} m{0.13\textwidth} m{0.45\textwidth}}
    \toprule
    \textbf{Error Type} & \textbf{Question} & \textbf{Answer} & \textbf{Self-Consistent Error} &  \textbf{Explanation} \\
    \midrule
    Pervasive Misconceptions &
    Which is the lightest of the widely used structural metals? 
    & Magnesium.
    & Aluminum.
    &  \textit{Misconceptions supporting this error can be found on many blogs and informal articles.} One article states, ``Aluminum is the lightest structural metal...'' Another blog: ``Aluminum vs Magnesium: The Lightest Structural Metals Compared.'' \\
    \midrule
    Easily Confused Concepts &
    When the earth is between the moon and the sun, what type of moon shows?
    & Full Moon.
    & A New Moon.
    &  \textit{These are two highly easily confused concepts.} A full moon occurs when the earth is between the moon and the sun. A new moon occurs when the moon is between the earth and the sun. \\
    \bottomrule
    \end{tabular}
\end{adjustbox}
\caption{Examples of Self-Consistent Errors.}
\label{tab:self_consistent_examples}
\end{table*}

\section{Self-Consistent Errors in LLMs}
\label{sec:consistent_error}

\subsection{Task Definition}
\textbf{Error detection} \citep{orgad2024llmsknowshowintrinsic,farquhar2024detecting}, also called hallucination detection,  seeks to decide whether an LLM's answer is factually correct.  
We use ``error detection'' due to the ambiguity of ``hallucination'' across domains \citep{Wang_Sennrich_2020_hallucination_definition}.
Starting from a QA dataset  
$
  \mathcal{Q}=\{(q_i,a_i)\}_{i=1}^{N},
$
where $q_i$ is a question and $a_i$ is its reference answer, we obtain the model's greedy response  
$
  r_i^{\mathrm{g}} = \mathcal{M}(q_i;\theta, T=0),
$
with language model $\mathcal{M}$ (parameters $\theta$) and temperature~$T$.  
Current work primarily targets greedy responses as they reflect the model’s best choice and facilitate reproducibility.
We label each prediction by comparing it with $a_i$, yielding $z_i\in\{0,1\}$ according to the procedure in Section~\ref{Experimental Settings}.  
This produces the error detection datasets  
$
  \mathcal{D}_\mathcal{M}=\{(q_i,r_i^{\mathrm{g}},z_i)\}_{i=1}^{N}.
$
At test time, the detector observes only $(q_i,r_i^{\mathrm{g}})$ and predict the error score $s_i = f(q_i, r_i^{\mathrm{g}})$.

\subsection{Definition of Self-Consistent Error}
\label{sec:definition_of_consistent_error}
We categorize errors as \textit{self-consistent} if the model repeatedly generates semantically equivalent incorrect responses across multiple stochastic samples for a given question, and as \textit{inconsistent} otherwise.
\begin{definition}[\textbf{Self-Consistent Error}]
\label{theorem_consistent_hallucination}
For a question $q_i$, we draw $k$ stochastic samples
\[
  r_{i,j}^{\mathrm{s}} = \mathcal{M}(q_i;\theta, T>0, j), \qquad j=1,\dots,k .
\]
If all samples are semantically equivalent to the greedy response,
\[
  r_{i,1}^{\mathrm{s}}\equiv r_{i,2}^{\mathrm{s}}\equiv\cdots\equiv r_{i,k}^{\mathrm{s}}\equiv r_i^{\mathrm{g}},
\]
and the greedy answer is judged incorrect ($z_i=0$), then $r_i^{\mathrm{g}}$ is a \emph{self-consistent error} for model $\mathcal{M}$.  
The relation $\equiv$ denotes semantic equivalence.
\end{definition}
 
To operationalize Definition~\ref{theorem_consistent_hallucination} and categorize errors in
$\mathcal{D}=\{(q_i,r_i^{\mathrm{g}},z_i)\}_{i=1}^{M}$, we proceed as follows.  
For every incorrect instance ($z_i=0$), we generate $k=15$ stochastic samples
$r_{i,1}^{\mathrm{s}},\dots,r_{i,15}^{\mathrm{s}}$ in addition to the greedy answer
$r_i^{\mathrm{g}}$.
We discuss the effect of $k$ in Appendix \ref{appendix:effec-of-k}.
Sampling is performed with temperature $T=0.5$, $\texttt{top\_p}=1$ and $\texttt{top\_k}=-1$, which is the commonly adopted settings in prior work \citep{semantic_entropy}.
Next, we test pairwise semantic equivalence within
$\{r_i^{\mathrm{g}}, r_{i,1}^{\mathrm{s}},\dots,r_{i,15}^{\mathrm{s}}\}$
with the NLI‑based criterion of \citet{semantic_entropy}, treating two responses as equivalent if they mutually entail each other.
An error $r_i^{\mathrm{g}}$ is labeled \textit{self-consistent} when all stochastic samples and greedy response are semantically
equivalent; otherwise, $r_i^{\mathrm{g}}$ is labeled \textit{inconsistent}.

\subsection{Analysis of Self-Consistent Errors}
\paragraph{Prevalence.} We investigate the prevalence of self-consistent errors across different model scales, including Qwen (Qwen2.5-3/7/14/32/72B-Instruct) and Llama (Llama3.2-1B/3B, 3.1-8/70B-Instruct)\footnote{As this work focuses on text-only models, we exclude vision LLMs (Llama3.2-11B/90B).}.
We use TriviaQA (TQA for short) \citep{joshi2017triviaqa} and SciQ \citep{sciq} datasets, which represent trivia and scientific knowledge domains, respectively. 
Figure~\ref{fig:intro} shows how the frequency of errors changes with model scale on SciQ, with TQA shown in Figure~\ref{fig:number_of_ce_appendix}.
Unlike inconsistent errors, which markedly decrease as models scale up, the number of self-consistent errors remains relatively stable, or even slightly increases.
This suggests that self-consistent errors, being more resistant to model scaling, will likely remain a persistent challenge, potentially becoming more concerning as LLMs continue to scale.
Therefore, analyzing and improving the capability to detect this class of errors becomes increasingly crucial.

\paragraph{Causes.} 
We provide case studies to reveal some of the possible underlying self-consistent errors. 
As shown in Table~\ref{tab:self_consistent_examples}, some self-consistent errors may originate from \textbf{pervasive misconceptions} on the internet, which may mix into the training data.
Others reflect LLMs’ consistent misunderstandings of \textbf{conceptual confusions} between closely related notions, such as ``New Moon'' and ``Full Moon'' in Table \ref{tab:self_consistent_examples}.
These insights are case-based, and a conclusive determination of the causes would require access to the full training data, which we leave to future work.

\begin{table*}[t]

    \centering
    \begin{adjustbox}{max width=\linewidth}
    \begin{tabular}{lcccccccccccc}
        \toprule
        \multirow{2.5}{*}{\textbf{Method}}&   \multicolumn{6}{c}{\textbf{Llama3.1-8b}} & \multicolumn{6}{c}{\textbf{Qwen2.5-7b}} \\
        \cmidrule(lr){2-7} \cmidrule(lr){8-13}
         & \textbf{SciQ-CE} & \textbf{SciQ-IE} & $\Delta\downarrow$  & \textbf{TQA-CE} & \textbf{TQA-IE} & $\Delta\downarrow$  & \textbf{SciQ-CE} & \textbf{SciQ-IE} & $\Delta\downarrow$ & \textbf{TQA-CE} & \textbf{TQA-IE} & $\Delta\downarrow$  \\
        \midrule
        Probability & 0.6325 & 0.8192 & 0.1867 & 0.6243 & 0.8455 & 0.2212 & 0.4571 & 0.6594 & 0.2023 & 0.5360 & 0.7148 & 0.1788 \\
        P(True) & 0.6251 & 0.7625 & 0.1374 & 0.6836 & 0.8018 & 0.1182 &  0.6158 & 0.7589 & 0.1431 & 0.7478 & 0.8373 & 0.0895\\
        SE & 0.4608 & 0.8820 & 0.4212 & 0.5216 & 0.9226 & 0.4010 & 0.4782 & 0.8247 & 0.3465& 0.4453 & 0.9119 & 0.4666 \\
        Probe (OOD) & 0.7287 & 0.9080 & 0.1793 & 0.7396 & 0.8989 & 0.1593 & 0.7487 & 0.8605 & 0.1118 & 0.7734 & 0.8911 & 0.1177  \\
        \rowcolor{gray!18} ~~+~cross-model & 0.8289 & 0.9385 & 0.1096 & 0.8024 & 0.9263 & 0.1239 & 0.8211 & 0.8893 & 0.0682 & 0.8691 & 0.9457 & 0.0766 \\
        Probe (ID) & 0.7917 & 0.9249 & 0.1332 & 0.7922 & 0.9272 & 0.1350 & 0.8250 & 0.8891 & \textbf{0.0641} & 0.8626 & 0.9467 & 0.0841 \\
        \rowcolor{gray!18} ~~+~cross-model & \textbf{0.8659} & \textbf{0.9408} & \textbf{0.0749} & \textbf{0.8470} & \textbf{0.9477} & \textbf{0.1007} & \textbf{0.8399} & \textbf{0.9078} & 0.0679 & \textbf{0.9088} & \textbf{0.9696} & \textbf{0.0608} \\
        \midrule
        \multirow{2.5}{*}{\textbf{Method}} &   \multicolumn{6}{c}{\textbf{Qwen2.5-14b}} & \multicolumn{6}{c}{\textbf{Mistral-12b}} \\
        \cmidrule(lr){2-7} \cmidrule(lr){8-13}
        & \textbf{SciQ-CE} & \textbf{SciQ-IE} & $\Delta\downarrow$ & \textbf{TQA-CE} & \textbf{TQA-IE} & $\Delta\downarrow$  & \textbf{SciQ-CE} & \textbf{SciQ-IE} & $\Delta\downarrow$ & \textbf{TQA-CE} & \textbf{TQA-IE} & $\Delta\downarrow$  \\
        \midrule
        Probability & 0.5480 & 0.7517 & 0.2037 & 0.4926 & 0.6477 & 0.1551 & 0.5858 & 0.7354 & 0.1496 & 0.6283 & 0.8605 & 0.2322 \\
        P(True) & 0.5287 & 0.6744 & 0.1457 & 0.7052 & 0.8515 & 0.1463 & 0.6595 & 0.7625 & 0.1030 & 0.7502 & 0.8545 & 0.1043 \\
        SE & 0.5427 & 0.8764 & 0.3337& 0.4425 & 0.9074 & 0.4649 & 0.3633 & 0.8210 & 0.4677 & 0.4494 & 0.9093 & 0.4599 \\
        Probe (OOD) & 0.7425 & 0.9025 & 0.1600 &  0.7871 &   0.9174 & 0.1303 & 0.7767 & 0.8553 & 0.0786 & 0.6927 & 0.8577 & 0.1650 \\
        \rowcolor{gray!18} ~~+~cross-model & 0.7927 & 0.9263 & 0.1336 & 0.8754 & 0.9115 & \textbf{0.0361} & 0.8458 & 0.9276 & 0.0818 & 0.7872 & 0.9069 & 0.1197 \\
        Probe (ID) & 0.7473 & 0.8582 & 0.1109 & 0.8512 & 0.9570 & 0.1058 & 0.7726 & 0.8652 & 0.0926& 0.8163 & 0.9063 &  0.0900 \\
        \rowcolor{gray!18} ~~+~cross-model & \textbf{0.8118} & \textbf{0.8931} & \textbf{0.0813}  & \textbf{0.9332} & \textbf{0.9776} & 0.0444 & \textbf{0.8548} & \textbf{0.9253} & \textbf{0.0705} & \textbf{0.8497} & \textbf{0.9359} & \textbf{0.0862} \\

        \bottomrule
    \end{tabular}
    \end{adjustbox}
    \caption{AUROC of error detection methods. $\Delta$ is the performance gap between CE and IE subsets. Cross-model methods use Qwen2.5-14B as the verifier except itself, for which we use Llama3.1-70b.}
    \label{tab:main_table}
\end{table*}

\section{How Well Do We Detect Self-Consistent Errors?}
\label{sec:how_well}
This section evaluates the performance of current error detection methods on self-consistent errors.

\subsection{Experiment Setup}
\label{Experimental Settings}
To ensure a fair comparison between two types of errors for supervised probe methods, we controlled the distribution of the dataset. 
We created specialized subsets for the two types of errors: \textbf{(i) CE subset}, containing only self-consistent errors as negative (incorrect) examples, and \textbf{(ii) IE subset}, containing only inconsistent errors as negative examples. 
Both subsets contain an identical number of negative examples and are paired with the same number of positive examples for training. 
This setup controls for the influence of training data volume on the supervised probe. 
The performance gap $\Delta$ between these two subsets reveals the different detection difficulty between the two types of errors.  

\textbf{Evaluation Metric.} Following prior works \citep{semantic_entropy,can_llm_express,duan2024shifting}, we evaluate error detection using the area under the receiver operator characteristic curve (\textbf{AUROC}).
We produce the correctness label $z_i$ by employing an LLM to evaluate whether the response is semantically equivalent to the ground truth answer, following \citep{just_ask_for_calibration,simpleqa}.
Details are provided in Appendix~\ref{sec:evaluation}.

\textbf{Baseline \& LLMs.} We evaluate four types of mainstream error detection methods on commonly used LLMs: Qwen2.5-7b/14b \citep{yang2024qwen2}, Llama3.1-8b, and Mistral-12b. Training-free baselines include: \textbf{(1)} \textbf{Probability} uses aggregated token probabilities \citep{orgad2024llmsknowshowintrinsic, factualconfidence, LN-Entropy}. \textbf{(2)} \textbf{P(True)} prompts LLM to self-critique correctness and uses the probability of ``True'' as the confidence score \citep{kadavath2022languagemodelsmostlyknow}. 
\textbf{(3)} \textbf{SE} \citep{semantic_entropy,farquhar2024detecting} samples multiple responses and calculates the entropy of their semantic clusters.
Supervised baselines include: \textbf{(4)} \textbf{Probe} that trains a simple feedforward neural network to detect error based on the hidden states of LLMs  \citep{lying}. 
We use the hidden states of the last token at the layer with the best validation performance.
We distinguish \textbf{Probe (ID)} (trained and evaluated on the same dataset) from \textbf{Probe (OOD)} (trained on one dataset, evaluated on another). For instance, Probe-OOD might be trained on the SciQ-CE before being evaluated on TQA-CE.
OOD evaluation is critical to ensure the probe captures truthfulness features, rather than overfitting to a single dataset \citep{orgad2024llmsknowshowintrinsic}.
Further details are in Appendix~\ref{sec:appendix_baseline_details}.

\subsection{Failures in Self-Consistent Errors}
As shown in \Cref{tab:main_table}, existing methods perform well on inconsistent errors (AUROC up to about 90\%). 
However, all methods suffer a substantial performance degradation on consistent errors.
SE, which performs best among training-free methods on IE subsets, exhibits the most dramatic decline on CE subsets, performing at or below random guessing.
This challenges the assumption that self-consistency implies correctness, revealing critical limitations in consistency-based detection methods.
Although supervised methods generally outperform training-free approaches on CE subsets, they still show significant performance degradation compared to IE subsets.
This indicates that self-consistent errors are more challenging to distinguish from correct responses even at the hidden state level. 
Furthermore, Probe (OOD) shows larger performance gaps ($\Delta$) compared to Probe (ID), suggesting that self-consistent errors are particularly difficult to detect when generalizing across different knowledge domains, TQA and SciQ. 

\section{Cross-Model Probe}

\begin{table}[t]
\centering
\begin{adjustbox}{max width=\linewidth}
\begin{tabular}{lrr}
\toprule
\multirow{2.5}{*}{\textbf{Verifier}} & \multicolumn{2}{c}{\textbf{Same Self-Consistent Errors}} \\
\cmidrule(lr){2-3}
& \textbf{ TQA-CE (4638)} & \textbf{SciQ-CE (952)} \\
\midrule
Qwen2.5-3b   & 311 ~~(6.7\%)  & 127 (13.3\%) \\
Qwen2.5-14b  & 639 (13.6\%)   & 273 (28.7\%) \\
Qwen2.5-72b  & 657 (14.2\%)   & 213 (22.4\%) \\
Llama3.2-3b  & 409 ~~(8.8\%)  & 194 (20.4\%) \\
Llama3.1-70b & 251 ~~(5.4\%)  & 151 (15.9\%) \\
\bottomrule
\end{tabular}
\end{adjustbox}
\caption{Number and percentage of questions in which the verifier produces a semantically equivalent self-consistent error to Qwen2.5-7B.}
\label{tab:model_specificity}
\end{table}

The poor performance of the evaluated methods on self-consistent errors suggests that features from the response-generating LLM alone may be insufficient for detecting such errors.
Fortunately, we observe that self-consistent errors are often model-specific and rarely overlap across different LLMs.
As shown in Table \ref{tab:model_specificity}, the highest overlap occurs between Qwen2.5-7B and Qwen2.5-14B, reaching only 28.7\%.
This observation motivates the use of an external verifier to supplement the detection of self-consistent errors.

Given the high efficiency \citep{mind} and strong performance of supervised probes, we build upon this approach.
Standard probe methods train a classifier to detect errors using internal states of $\mathcal{M}$ which generate the response $r_i^{\mathrm{g}}$:
$$
s_{i}^{\mathcal{M}} = \mathrm{Probe}_{\mathcal{M}}(\mathbf{h}_i^{\mathcal{M}}), \quad \mathbf{h}_{i}^{\mathcal{M}} = \phi_{\mathcal{M}}^{(l,t)}\bigl(q_i,\, r_i^{\mathrm{g}}\bigr)
$$
where $\phi_{\mathcal{M}}^{(l,t)}$ extracts internal states from layer $l$ and token position $t$ of model $\mathcal{M}$.
We introduce a cross-model probe that leverages an external verifier LLM $\mathcal{V}$ to embed the responses generated by $\mathcal{M}$ and trains a separate $\mathrm{Probe}_{\mathcal{V}}$:
$$
s_{i}^{\mathcal{V}} = \mathrm{Probe}_{\mathcal{V}}(\mathbf{h}_i^{\mathcal{V}}), \quad \mathbf{h}_{i}^{\mathcal{V}} = \phi_{\mathcal{V}}^{(l,t)}\bigl(q_i,\, r_i^{\mathrm{g}}\bigr)
$$

The final error score combines both probes through an integration parameter $\lambda$:
$$
\mathrm{score}_i = (1 - \lambda) \cdot s_{i}^{\mathcal{M}} + \lambda \cdot s_{i}^{\mathcal{V}}
$$

\subsection{Effectiveness of Cross-Model Probe}

\paragraph{Implementation Details.} We select Qwen2.5-14B as the verifier for all other models except itself, for which we use Llama3.1-70b. 
$\lambda$ is selected from $\{0, 0.05, 0.1, \dots, 1.0\}$ by choosing the value that yields the best validation performance.

Building on the experiment settings in Section \ref{Experimental Settings}, we evaluate the proposed cross-model probe and show the result in \Cref{tab:main_table}.
Overall, the \textbf{cross-model probe demonstrates significant performance improvements on CE subsets}, regardless of in-domain or out-of-domain settings, highlighting its effectiveness in addressing the challenges of self-consistent errors.
Notably, this improvement does not sacrifice performance on inconsistent errors, where our method also achieves slight gains.
We further analyze the impact of verifier selection and the integration parameter~$\lambda$:

\paragraph{Effect of Different Verifiers.} 
We fix the original LLM as Qwen2.5-7b and select several verifiers from different series and scales to study the impact of different verifiers.
Overall, \Cref{tab:verifier-results} demonstrates that \textbf{our method outperforms the best baseline across all verifiers}, including the smallest 3B model.
Our empirical results also suggest: (i) verifiers from different model series outperform same-series ones, and (ii) larger-scale verifiers perform better than smaller-scale counterparts.

\begin{figure}[t]
    \centering
    \begin{adjustbox}{width=0.97\columnwidth}
        \includegraphics{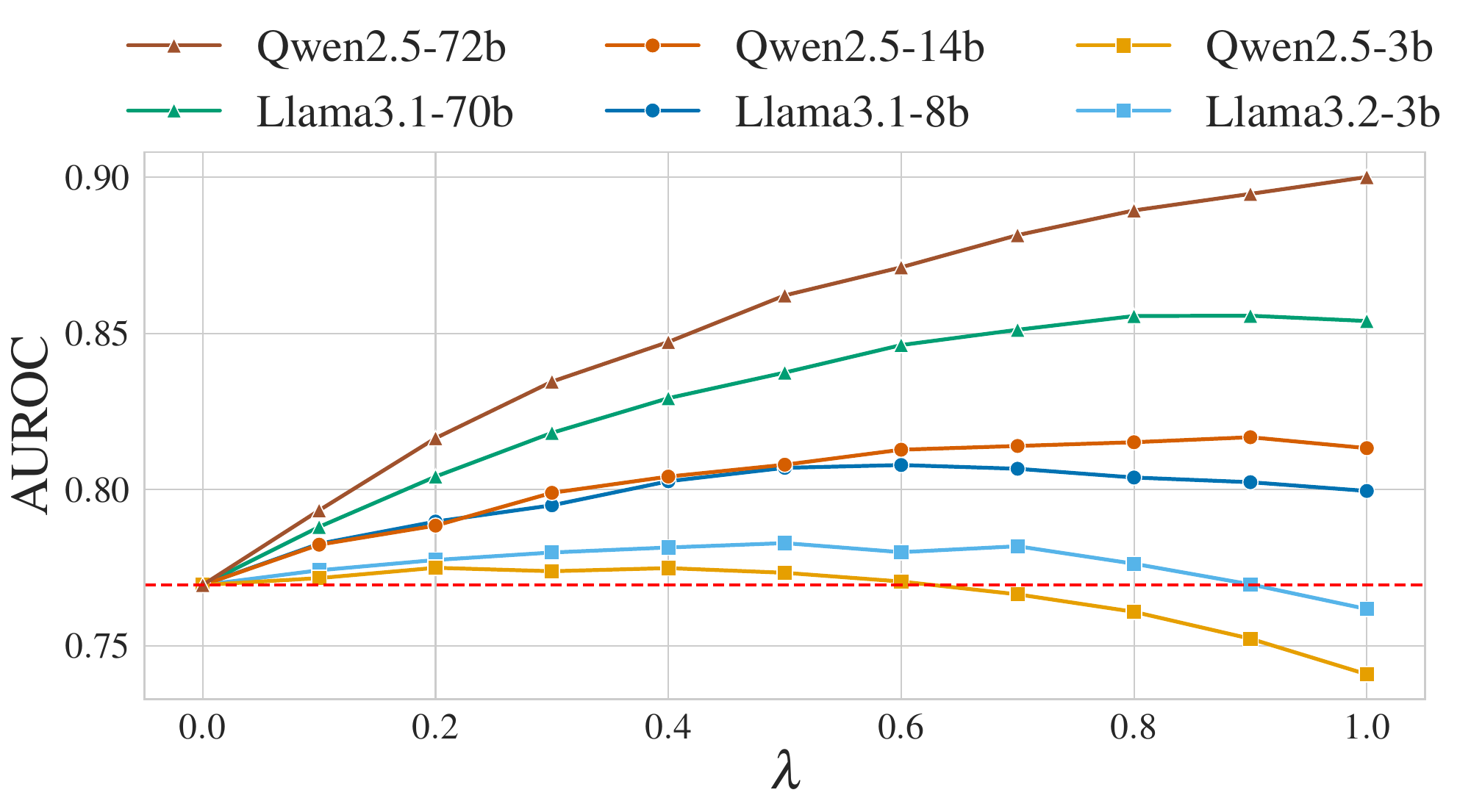}
    \end{adjustbox}
    \caption{Performance of cross-model under different $\lambda$ values with various verifiers. The \textcolor{red}{red} dashed line indicates the best baseline probe.}
    \label{fig:lambda}
\end{figure}

\paragraph{Effect of $\lambda$.} We conduct an analysis of $\lambda\in[0,1]$ on the SciQ-CE validation set using Qwen2.5-7B as the response LLM. 
(i) Overall, \textbf{our method consistently outperforms the baseline across a broad range of $\lambda$ values}, as shown in \Cref{fig:lambda}. 
Even with a weaker verifier Qwen2.5-3B, our method still outperforms the baseline over a wide $\lambda$ range ($0 < \lambda \leq 0.6$).
(ii) Simply fixing $\lambda=0.5$ provides significant gains across all tested verifiers, which makes deployment easy without extensive tuning.
(iii) We also provide empirical guidance on selecting a better $\lambda$.
Larger verifiers prefer higher $\lambda$, while smaller verifiers tend to perform better with relatively lower values (e.g., $\lambda \in [0.2, 0.5]$).

\begin{table}[t]
\centering
\begin{tabular}{@{}m{0.84\linewidth} m{0.13\linewidth}@{}}
    \begin{adjustbox}{max width=\linewidth}
        \begin{tabular}{lrrrr}
        \toprule
        \textbf{Method} & \textbf{SciQ-CE} & \textbf{SciQ-IE} & \textbf{TQA-CE} & \textbf{TQA-IE} \\
        \midrule
        Probe (Qwen2.5-7B) & \numinc{0.8250}{0} & \numinc{0.8786}{0} & \numinc{0.8662}{0} & \numinc{0.9468}{0} \\
        + Qwen2.5-3B       & \numinc{0.8357}{0.013} & \numinc{0.8834}{0.005} & \numinc{0.8712}{0.006} & \numinc{0.9495}{0.003} \\
        + Llama3.2-3B      & \numinc{0.8453}{0.025} & \numinc{0.8851}{0.007} & \numinc{0.8828}{0.019} & \numinc{0.9569}{0.011} \\
        + Qwen2.5-72B      & \numinc{0.8689}{0.053} & \numinc{0.9290}{0.057} & \numinc{0.9377}{0.083} & \numinc{0.9815}{0.037} \\
        \textbf{+ Llama3.1-70B} &
        \textbf{\numinc{0.8794}{0.066}} & \textbf{\numinc{0.9353}{0.065}} &
        \textbf{\numinc{0.9511}{0.098}} & \textbf{\numinc{0.9852}{0.041}} \\
        \bottomrule
        \end{tabular}
    \end{adjustbox}
& %
    \begin{adjustbox}{valign=m}
    \colorbarverticalinc[2]
  \end{adjustbox} \\ %
\end{tabular}

\caption{Cross-model probe with different verifiers. Shading indicates relative improvement over the baseline probe.}
\label{tab:verifier-results}
\end{table}

\section{Related work}

In recent years, error detection has attracted increasing attention as LLMs are deployed widely \citep{inside,farquhar2024detecting, haloscope}.
A prominent line of work is the consistency-based method \citep{selfcheckgpt,semantic_entropy,generating_with_confidence,can_llm_express, SAC3, selfconsistency}, which leverages the semantic agreement between multiple sampled responses, implicitly treating consistency as a proxy for correctness.
In addition, other signals from the LLM itself are also employed to detect errors, such as sequence probability \citep{LN-Entropy,duan2024shifting} and verbalized confidence \citep{kadavath2022languagemodelsmostlyknow, just_ask_for_calibration,lin2022teaching}.
Beyond these training-free methods, supervised probes further leverage LLMs' internal states and achieve strong performance \citep{lying,li2023inferencetime, InternalInspector, pollmgraph, geometry, burns2023discovering, factualconfidence}.
However, these methods are constrained by their reliance on the model’s own signals, which are insufficient for detecting self-consistent errors.
We demonstrate that leveraging hidden states from external models can help overcome these limitations, offering a new direction for error detection.
A more detailed review is provided in Appendix \ref{sec:appendix:related_work}.

\section{Conclusion}
This work reveals that current error detection methods struggle to detect \textit{self-consistent errors}, where the LLM repeatedly produces the same incorrect response across multiple stochastic samples.
We demonstrate that this problem is critical because it persists under scaling laws, i.e., the frequency of such errors remains constant or even increases as the model size grows.
To address this, we propose a cross-model probe that incorporates hidden states from external models.
The effectiveness of this approach highlights a promising new direction for more reliable error detection.
While this work primarily focuses on detecting self-consistent errors, it also raises critical research questions about their root causes and mitigation strategies, calling on the community for further exploration.

\section{Limitations}
The underlying causes of consistent errors still require deeper investigation. 
These systematic failures may stem from prevalent misconceptions in training data or biases introduced during the supervised training phase.
Future work may construct controlled experiments to investigate the causes.

\section{Ethics Statement}

\paragraph{Data} All data used in this study are publicly available and do not pose any privacy concerns.

\paragraph{AI Writing Assistance} In our study, we only employed ChatGPT to polish our textual expressions
rather than to generate new ideas or suggestions.

\section*{Acknowledgments}

This work was supported by the National Key
R\&D Program of China (2022YFB3103700, 2022YFB3103704), the Strategic Priority Research Program of the Chinese Academy of Sciences (XDB0680201), the Beijing Natural Science Foundation (4252023), the Innovation Funding of ICT, CAS (E361120), the National Natural Science Foundation of China (No.62172393), and
Major Public Welfare Project of Henan Province
(No.201300311200).

\bibliography{custom}

\appendix

\section{Appendix}
\label{sec:appendix}

\subsection{The Number of Consistent and Inconsistent Errors}
\begin{figure*}
    \centering
    \includegraphics[width=0.8\textwidth]{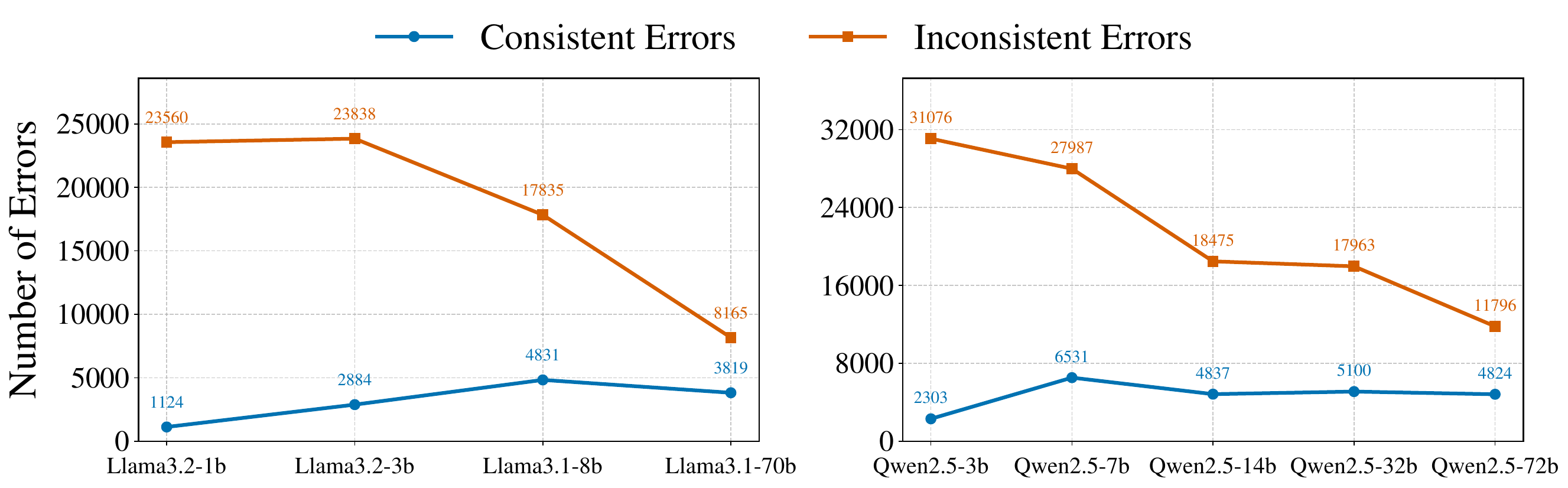}
    \caption{The number of self-consistent and inconsistent errors across different scales of LLMs on TriviaQA.}
    \label{fig:number_of_ce_appendix}
\end{figure*}

\Cref{fig:number_of_ce_appendix} shows the number of consistent and inconsistent errors for different LLMs.

\begin{figure*}
    \centering
    \includegraphics[width=0.80\linewidth]{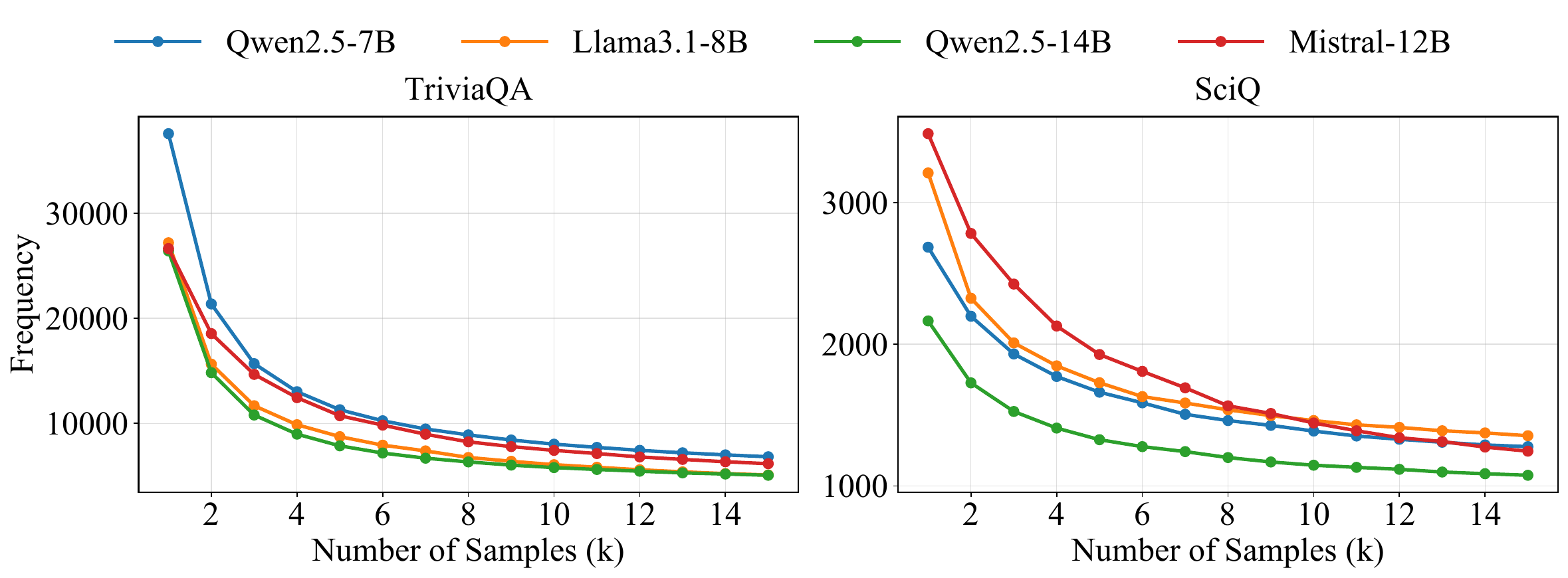}
    \caption{The number of self-consistent errors with different sample numbers $k$.}
    \label{fig:effect_of_k}
\end{figure*}

\subsection{Related Work}
\label{sec:appendix:related_work}

\textbf{Error Detection.}
Large language models (LLMs) often generate responses that appear plausible but contain factual inaccuracies. 
This challenge underscores the critical importance of accurately detecting errors in LLM-generated content for establishing trustworthiness. 
While this task is also referred to as ``hallucination detection'' \citep{inside, farquhar2024detecting, haloscope}, we adopt the term ``error detection'' to avoid ambiguity, as ``hallucination'' carries domain-specific meanings across different fields \citep{Huang2025hallucinationdefinition, Wang_Sennrich_2020_hallucination_definition}.

\textbf{Training-Free Error Detection.}
A prominent approach to error detection involves estimating the uncertainty inherent in the model itself. 
Methods in this category include analyzing response probabilities \citep{LN-Entropy, duan2024shifting} and eliciting verbalized confidence scores directly from the model \citep{just_ask_for_calibration, lin2022teaching,can_llm_express}. 
Among these methods, consistency-based uncertainty \citep{selfcheckgpt,semantic_entropy,generating_with_confidence,can_llm_express,inside,SAC3, inside} has received considerable attention.
Building on the assumption that consistent responses are more likely to be factually correct \citep{selfconsistency}, consistency-based methods sample multiple responses and compute semantic consistency among them to detect hallucinations.

\textbf{Supervised Probe.}
In contrast to the above methods, probe-based approaches employ supervised learning to identify truthfulness features embedded within LLMs' internal states. 
Several previous works \citep{geometry, lying, burns2023discovering, li2023inferencetime, inside} have claimed that there existed truthfulness features in the internal states of LLMs.
Based on the assumption, numerous studies have tried to detect hallucination using the features from LLMs' own internal states \citep{kadavath2022languagemodelsmostlyknow, lying, InternalInspector, pollmgraph}.
These works trained a probe, a simple classifier, to predict whether the response of LLMs is correct based on the internal states.
As the probe is often a simple multi-layer perceptron, these methods need very low computation cost both during inference time and the training process \citep{mind}.
Moreover, recent comparative studies \citep{factualconfidence} have demonstrated their superior performance over other consistency-based, probability-based, and verbalized methods.

\textbf{Self-Consistent Error.}
Prior consistency-based error detectors \citep{farquhar2024detecting, SAC3, inside} also acknowledged the limitations of consistency-based methods in handling self-consistent errors.
However, they neither quantify the extent of performance degradation nor systematically examine the prevalence of such errors.
Moreover, their analysis is limited to consistency-based methods, leaving the effect on other types of methods unclear.
In contrast, our work provides a comprehensive evaluation across four mainstream categories of error detection methods and reveals that self-consistent errors pose a universal challenge, leading to significant performance drops across all methods, not just those relying on sample consistency.

\textbf{Cross-Model Checking.}
\citet{SAC3} and concurrent work \citep{xue2025verify} propose to detect errors by sampling multiple responses from both the target model and an external model, followed by measuring their agreement.
However, these approaches require 10–20 additional generations per query across both models, making them impractical for real-time usage.
In contrast, our Cross-Model Probe offers a novel and efficient alternative that requires only a single forward pass through a verifier model.
Furthermore, our empirical analysis provides valuable insights for verifier selection, extending the understanding beyond previous approaches.

\subsection{Baseline Method Implementation Details}
\label{sec:appendix_baseline_details}

Here we provide detailed implementation details for the baseline error detection methods evaluated in Section~\ref{Experimental Settings}.

\textbf{(i)} \textbf{Probability}: Several studies have employed the aggregated token probabilities to detect errors \citep{orgad2024llmsknowshowintrinsic, factualconfidence, LN-Entropy}. Following prior work \citep{orgad2024llmsknowshowintrinsic}, we average the log-probabilities of all generated tokens in a response. This average log-probability serves as the error detection indicator, where lower values suggest a higher likelihood of error.

\textbf{(ii)} \textbf{P(True)}: This method follows the prompting strategy introduced by \citet{kadavath2022languagemodelsmostlyknow}, where the LLM is directly queried to assess the correctness of its own output. Specifically, we construct the following prompt:
\begin{quote}
    \ttfamily
    Question: \{question\} \\
    Possible answer: \{response\} \\
    Is the possible answer: \\A. True \quad B. False \\
    The possible answer is:
\end{quote}
The model's confidence is then quantified as the probability it assigns to the token ``A''. A higher probability indicates greater model confidence in the correctness of its response.

\textbf{(iii)} \textbf{SE} (Semantic Entropy): As proposed by \citet{semantic_entropy} and further explored by \citet{farquhar2024detecting}, semantic entropy estimates uncertainty over the meaning conveyed by a response, rather than just the token sequence. 
Higher semantic entropy suggests greater uncertainty about the response's meaning and thus a higher likelihood of error.
Following the implementation details recommended by \citet{semantic_entropy}, we set the sampling parameters as follows: temperature 0.5, number of samples 10, top\_p = 1.0, and top\_k = -1.

\textbf{(iv)} \textbf{Probe}:
Following \citet{lying}, we implement a probe using a three-layer feedforward neural network (FFN) with ReLU activations and hidden dimensions set to (256, 128, 64). 
The model is trained with cross-entropy loss.
To select the most informative hidden layer, we train a separate probe on the output of each layer and choose the one that achieves the highest AUROC on the validation set. To mitigate overfitting, the probe is trained for a fixed number of epochs, and we select the checkpoint with the best validation performance for final evaluation.

\subsection{Evaluation Metric}
\label{sec:evaluation}
Following prior works
\citep{semantic_entropy,can_llm_express,duan2024shifting}, we evaluate error detection using the area under the receiver operator characteristic curve (\textbf{AUROC}), which reflects models' ability to distinguish incorrect and correct responses.
We produce the correctness label $z_i$ by employing an LLM to evaluate whether the response is semantically equivalent to the ground truth answer, following \citep{just_ask_for_calibration,simpleqa}.
To ensure reproducibility, we employ the powerful open-source model, Llama-3.1-70b.
Inspired by \citep{simpleqa}, we use the prompt in \Cref{lst:evaluation_prompt} to check the correctness of the generated response.
This prompt categorizes responses into correct, incorrect, and refusal. 
In our experiments, we filter out the refusal responses, as our focus is on effectively distinguishing between correct and incorrect responses.
A manual review finds that only 1 out of 300 samples disagrees with human annotation, demonstrating the reliability of the correctness label.

\subsection{Effect of $k$}
\label{appendix:effec-of-k}

\Cref{fig:effect_of_k} illustrates how the number of self-consistent errors changes as the number of stochastic samples $k$ increases from 1 to 15 on TriviaQA (left) and SciQ (right). 
In both datasets, we observe a steep initial decline, followed by a convergence trend when $k > 10$.
We set $k=15$ in all subsequent experiments, as it offers a statistically stable estimate of the self-consistent error rate while maintaining acceptable computational cost.

\subsection{Prompt}
\onecolumn
\begin{tcolorbox}[
  colback=gray!5,
  colframe=gray!75!black,
  title=Evaluation Prompt,
  fonttitle=\bfseries,
  breakable,
  width=\textwidth,
  left=3pt,
  right=3pt,
  top=3pt,
  bottom=3pt,
  boxsep=2pt,
  label={lst:evaluation_prompt}
]
{\footnotesize
\begin{lstlisting}[breaklines=true, columns=flexible]
Your job is to look at a question, some gold targets, and a predicted answer, and then assign a grade of either ["CORRECT", "INCORRECT", "NOT_ATTEMPTED"].
First, I will give examples of each grade, and then you will grade a new example.

The following are examples of CORRECT predicted answers.

Question: What are the names of Barack Obama's children?
Gold target: ["Malia Obama and Sasha Obama", "malia and sasha"]
Predicted answer 1: sasha and malia obama
Predicted answer 2: most people would say Malia and Sasha, but I'm not sure and would have to double check
Predicted answer 3: Barack Obama has two daughters. Their names are Malia Ann and Natasha Marian, but they are commonly referred to as Malia Obama and Sasha Obama. Malia was born on July 4, 1998, and Sasha was born on June 10, 2001.

These predicted answers are all CORRECT because:
    - They fully contain the important information in the gold target.
    - They do not contain any information that contradicts the gold target.
    - Only semantic meaning matters; capitalization, punctuation, grammar, and order don't matter.
    - Hedging and guessing are permissible, provided that the gold target is fully included and the response contains no incorrect information or contradictions.


The following are examples of INCORRECT predicted answers.

Question: What are the names of Barack Obama's children?
Gold target: ["Malia and Sasha"]
Predicted answer 1: Malia.
Predicted answer 2: Malia, Sasha, and Susan.
Predicted answer 3: Barack Obama does not have any children.
Predicted answer 4: I think it's either Malia and Sasha. Or it could be Malia and Jackie. Or it could be Joey and Malia.
Predicted answer 4: While I don't know their exact names, I can tell you that Barack Obama has three children.
Predicted answer 5: It's possible you may mean Betsy and Olivia. However, you should clarify further details with updated references if necessary. Is that the correct answer?
Predicted answer 6: It may be the case that Obama's child is named James. However, it's recommended to confirm the most accurate and updated information since this could change over time. This model may not always reflect the most current information.

These predicted answers are all INCORRECT because:
    - A factual statement in the answer contradicts the gold target. Incorrect statements that have some hedging (e.g., "it is possible that", "although i'm not sure, i think") are also considered incorrect.


The following are examples of NOT_ATTEMPTED predicted answers.

Question: What are the names of Barack Obama's children?
Gold target: ["Malia and Sasha"]
Predicted answer 1: I don't know.
Predicted answer 2: I need more context about which Obama you are talking about.
Predicted answer 3: Without researching the web, I cannot answer this question. However, I can tell you that Barack Obama has two children.
Predicted answer 4: Barack Obama has two children. I know that one of them is Malia, but I'm not sure about the other one.

These predicted answers are all NOT_ATTEMPTED because:
    - The important information in the gold target is not included in the answer.
    - No statements in the answer contradict the gold target.


Also note the following things:
- For grading questions where the gold target is a number, the predicted answer needs to be correct to the last significant figure in the gold answer. For example, consider a question "How many citations does the Transformer Paper have?" with gold target "120k". 
    - Predicted answers "120k", "124k", and 115k" are all CORRECT. 
    - Predicted answers "100k" and "113k" are INCORRECT. 
    - Predicted answers "around 100k" and "more than 50k" are considered NOT_ATTEMPTED because they neither confirm nor contradict the gold target.
- The gold target may contain more information than the question. In such cases, the predicted answer only needs to contain the information that is in the question.
    - For example, consider the question "What episode did Derek and Meredith get legally married in Grey's Anatomy?" with gold target "Season 7, Episode 20: White Wedding". Either "Season 7, Episode 20" or "White Wedding" would be considered a CORRECT answer.
- Do not punish predicted answers if they omit information that would be clearly inferred from the question.
    - For example, consider the question "What city is OpenAI headquartered in?" and the gold target "San Francisco, California". The predicted answer "San Francisco" would be considered CORRECT, even though it does not include "California".
    - Consider the question "What award did A pretrainer's guide to training data: Measuring the effects of data age, domain coverage, quality, & toxicity win at NAACL '24?", the gold target is "Outstanding Paper Award". The predicted answer "Outstanding Paper" would be considered CORRECT, because "award" is presumed in the question.
    - For the question "What is the height of Jason Wei in meters?", the gold target is "1.73 m". The predicted answer "1.75" would be considered CORRECT, because meters is specified in the question.
    - For the question "What is the name of Barack Obama's wife?", the gold target is "Michelle Obama". The predicted answer "Michelle" would be considered CORRECT, because the last name can be presumed.
- Do not punish for typos in people's name if it's clearly the same name. 
    - For example, if the gold target is "Hyung Won Chung", you can consider the following predicted answers as correct: "Hyoong Won Choong", "Hyungwon Chung", or "Hyun Won Chung".


Here is a new example. Simply reply with either CORRECT, INCORRECT, NOT_ATTEMPTED. Don't apologize or correct yourself if there was a mistake; we are just trying to grade the answer.

Question: {question}
Gold target: {target}
Predicted answer: {predicted_answer}

Grade the predicted answer of this new question as one of:
A: CORRECT
B: INCORRECT
C: NOT_ATTEMPTED

Just return the letters "A", "B", or "C", with no text around it.
\end{lstlisting}
}
\end{tcolorbox}
\twocolumn

\end{document}